\documentclass{article}
\usepackage[utf8]{inputenc}
\usepackage{geometry}
\usepackage{amsmath, amssymb, amsthm}
\usepackage{graphicx}
\usepackage{hyperref}
\usepackage{listings}
\usepackage{booktabs}
\usepackage{color}
\usepackage{float}

\title{\textbf{Valori: A Deterministic Memory Substrate for AI Systems}}
\author{Varshith Gudur \\ 
\texttt{varshith.gudur17@gmail.com} \\ 
\textit{Valori Kernel Project} \\ 
\textit{Independent Researcher}}
\date{\today}

\begin{document}

\maketitle

\begin{abstract}
Modern AI systems rely on vector embeddings stored and searched using floating-point arithmetic. While effective for approximate similarity search, this design introduces fundamental non-determinism: identical models, inputs, and code can produce different memory states and retrieval results across hardware architectures (e.g., x86 vs. ARM). This prevents replayability and safe deployment, leading to silent data divergence that prevents post-hoc verification and compromises audit trails in regulated sectors. We present \textbf{Valori}, a deterministic AI memory substrate that replaces floating-point memory operations with fixed-point arithmetic (Q16.16) and models memory as a replayable state machine. Valori guarantees bit-identical memory states, snapshots, and search results across platforms. We demonstrate that non-determinism arises before indexing or retrieval and show how Valori enforces determinism at the memory boundary. Our results suggest that deterministic memory is a necessary primitive for trustworthy AI systems. The reference implementation is open-source and available at \url{https://github.com/varshith-Git/Valori-Kernel} (archived at \url{https://zenodo.org/records/18022660}).
\end{abstract}


\section{Introduction}
Artificial Intelligence systems increasingly rely on long-term memory to retrieve context, ground generation (RAG), and maintain state in agentic workflows. In these systems, memory is assumed to be a stable foundation: if an agent stores a fact today, it should be retrievable tomorrow, regardless of component upgrades or server migrations. 

However, in reality, modern AI memory is not replayable. The reliance on standard floating-point arithmetic (IEEE 754) for storing and indexing vector embeddings introduces subtle, hardware-dependent variations. A vector computed and stored on an x86 server may have a different bit-level representation than the supposedly "identical" vector on an ARM-based edge device. These discrepancies propagate through the indexing and retrieval pipeline, leading to divergent search results and non-reproducible system behavior.

This non-determinism is not widely acknowledged as a critical failure mode, often being dismissed as negligible floating-point error. Yet, in safety-critical applications—such as robotics, defense, and financial auditing—bit-level reproducibility is a correctness requirement, not an optional feature.

This paper argues that determinism must be enforced at the AI memory layer. We introduce \textbf{Valori}, a memory kernel designed to provide a mathematically rigorous, hardware-agnostic foundation for AI state.

\section{Background \& Motivation}

The root cause of memory non-determinism lies in the nature of floating-point arithmetic itself.

\subsection{Floating-Point Non-Determinism}
The IEEE 754 standard for floating-point arithmetic allows for implementation-defined behavior in operations such as Fused Multiply-Add (FMA) and the ordering of parallel reductions (e.g., SIMD instructions). Compilers and hardware architectures (x86\_64 with AVX vs. ARM64 with NEON) optimize these operations differently.
\\
\\
\textbf{FMA}: $a \times b + c$ can be computed with a single rounding step (FMA) or two (Multiply then Add). These yield slightly different results.

\vspace{0.5em}
\noindent\textbf{Associativity}: Floating-point addition is not associative ($ (a+b)+c \neq a+(b+c) $). Parallel reduction strategies change the order of operations, altering the final sum.

\vspace{0.5em}
\noindent\textbf{SIMD and Auto-Vectorization}: Compilers often auto-vectorize reduction loops using hardware-specific instructions (e.g., AVX2, AVX-512, NEON). These instruction sets have different register widths and association orders, causing the biggest source of non-determinism in vector databases today even when source code is identical.

\subsection{The Embedding Pipeline}
In a typical RAG or agent pipeline, a neural network (e.g., a Transformer) generates a high-dimensional vector (embedding) from an input text. This vector is assumed to be the "semantic address" of the data. However, because inference runs on floating-point hardware, non-determinism enters the system at embedding generation, before indexing or retrieval. As we show in Section \ref{sec:evidence}, the same model on different chips produces different raw bits for the same input.

\section{Problem Definition}

We define the requirements for a trustworthy memory system using the formalism of a state machine.

\subsection{Memory State and Replayability}
Let $S_t$ be the state of the memory system at logical time $t$. Let $C_t$ be a command applied to the memory (e.g., `insert(vec)`, `delete(id)`, `link(a, b)`). The transition function $F$ defines the next state:
$$ S_{t+1} = F(S_t, C_t) $$

\textbf{Deterministic Memory} requires that for any valid initial state $S_0$ and any sequence of commands $C_1, ..., C_N$, the final state $S_N$ is identical regardless of the computing environment (CPU architecture, OS, compiler version).
$$ \forall \text{Env}_A, \text{Env}_B : \text{Apply}(S_0, \{C_i\})|_A \equiv \text{Apply}(S_0, \{C_i\})|_B $$

Typical vector databases using `f32` violate this property because the transition function $F$ (specifically distance calculations and index updates) depends on hardware-specific floating-point behavior.

\section{Empirical Evidence of Non-Determinism}
\label{sec:evidence}

To quantify the divergence, we conducted a controlled experiment generating embeddings on two standard developer machines:
\begin{enumerate}
    \item \textbf{x86}: Windows PC, x86\_64 architecture.
    \item \textbf{ARM}: MacBook Pro, ARM64 (Apple Silicon).
\end{enumerate}

Both machines ran the exact same Python code, using `sentence-transformers` (v2.6.1) to encode a standard input text list. The scripts used to reproduce this experiment are provided below.

\subsection{Reproduction Setup}

First, we generate embeddings for a fixed set of sentences. This script uses a deterministic seed for the model initialization (handled by the library), but the floating-point interference occurs during the forward pass.

\begin{lstlisting}[
    language=Python, 
    caption=Generation Script (embed.py), 
    basicstyle=\small\ttfamily, 
    breaklines=true,
    frame=single 
]
import numpy as np
from sentence_transformers import SentenceTransformer

model = SentenceTransformer("sentence-transformers/all-MiniLM-L6-v2")

texts = [
    "Revenue for April",
    "What is the profit in April?",
    "April financial summary",
    "Total earnings last month",
    "Completely unrelated sentence"
]

embeddings = model.encode(texts, normalize_embeddings=False)
np.save("embeddings.npy", embeddings)
\end{lstlisting}

Next, we inspect the raw hexadecimal representation of the floating-point values to detect bit-level differences that standard printing would hide.

\begin{lstlisting}[
    language=Python, 
    caption=Inspection Script (inspect.py), 
    basicstyle=\small\ttfamily, 
    breaklines=true,
    frame=single 
]
import numpy as np

emb = np.load("embeddings.npy")

# Print the hex representation of the first 5 floats of vector 0
for i in range(5):
    print(hex(emb[0][i].view(np.uint32)))
\end{lstlisting}

\subsection{Results}

We inspected the raw bits of the generated floating-point embeddings produced by the scripts above. The results are shown in Table \ref{tab:divergence}.

\begin{table}[ht]
\centering
\caption{Bit-Level Divergence of Identical Embeddings (First 5 Dimensions)}
\label{tab:divergence}
\begin{tabular}{ccc}
\toprule
Dimension & x86 Value (Hex) & ARM Value (Hex) \\
\midrule
0 & \texttt{0xbd8276f8} & \texttt{0xbd8276fc} \\
1 & \texttt{0x3d6bb481} & \texttt{0x3d6bb470} \\
2 & \texttt{0x3d1dcdf1} & \texttt{0x3d1dcdf9} \\
3 & \texttt{0xbd601d21} & \texttt{0xbd601d16} \\
4 & \texttt{0x3b761ffb} & \texttt{0x3b762229} \\
\bottomrule
\end{tabular}
\end{table}

As shown in Table \ref{tab:divergence}, the binary representation differs in every single dimension examined. While the cosine similarity between these divergent vectors remains extremely high ($>0.9999$), the bit-level difference breaks state replicability guarantees. This confirms that memory state acts as a "forking path" from the moment of creation. These discrepancies alter distance rankings, as the L2 distance between a query and a document will vary slightly, potentially changing the set of retrieved results ($k$-NN) and causing the agent's behavior to diverge depending on the server it runs on.

\section{Valori Design}

Valori solves this problem by enforcing a strict determinism boundary. \textbf{Valori does not attempt to make neural inference deterministic; instead, it defines a strict boundary at which non-deterministic model outputs are normalized into a deterministic memory state.} It is architected as a \texttt{no\_std} Rust kernel that can run on bare metal, WASM, or standard operating systems.

\subsection{Fixed-Point Arithmetic (Q16.16)}
Valori replaces all `f32/f64` operations with Q16.16 fixed-point arithmetic as the initial default.
\begin{itemize}
    \item \textbf{Format}: 32-bit signed integers, where the lower 16 bits represent the fractional part. Range: $[-32768, 32767]$. Resolution: $ \approx 0.000015 $.
    \item \textbf{Rationale}: Q16.16 was selected for the reference implementation because it offers a balance of efficient execution on 32-bit embedded MCUs and sufficient precision for normalized embeddings (typically $[-1, 1]$). It is not a hard logical limit; the kernel design supports higher precision contracts (e.g., Q32.32) for future enterprise applications.
    \item \textbf{Operations}: Addition and subtraction are integer ops. Accumulators use i64 (or wider) intermediates during the dot product summation to prevent overflow before narrowing back to the stored Q16.16 format.
    \item \textbf{Determinism}: Since Q16.16 relies on standard integer ALU instructions (which are consistent across architectures), numerical results are guaranteed bit-identical on x86, ARM, RISC-V, and WASM.
\end{itemize}

\subsection{Memory as a State Machine}
The kernel is a pure state machine implemented without standard library IO \texttt{no\_std}.
\begin{itemize}
    \item \textbf{Inputs}: Commands (Insert, Link, Delete) must be serialized and deterministic.
    \item \textbf{State}: The `Kernel` struct encapsulates all vector data, graph selection, and metadata.
    \item \textbf{Snapshot/Restore}: Because the state is deterministic, the entire memory can be serialized to a snapshot file. Restoring this snapshot on a different machine guarantees an exact replica of the memory state, down to the last bit.
\end{itemize}

\subsection{Architecture Separation}
The Valori Kernel acts as a deterministic execution environment for memory operations. All external inputs—whether originating from Python, HTTP clients, or distributed nodes—are normalized at the kernel boundary into a fixed-point representation with a well-defined precision contract (e.g., Q16.16). Once inside the kernel, memory updates are applied through a pure state-machine transition function, ensuring that identical command sequences produce bit-identical memory states across architectures. Index structures, snapshotting, and restoration are implemented entirely within this deterministic domain.

The system is logically divided into:
\begin{itemize}
    \item \textbf{Kernel \texttt{no\_std}}: The core logic. Validates inputs, converts floats to fixed-point immediately at the boundary, and performs all indexing math.
    \item \textbf{Node (`std`)}: An outer layer (Axum/Tokio) that provides HTTP APIs, persistence, and networking. It wraps the kernel but does not alter its logic.
\end{itemize}

\begin{figure}[ht]
    \centering
    \includegraphics[width=0.8\textwidth]{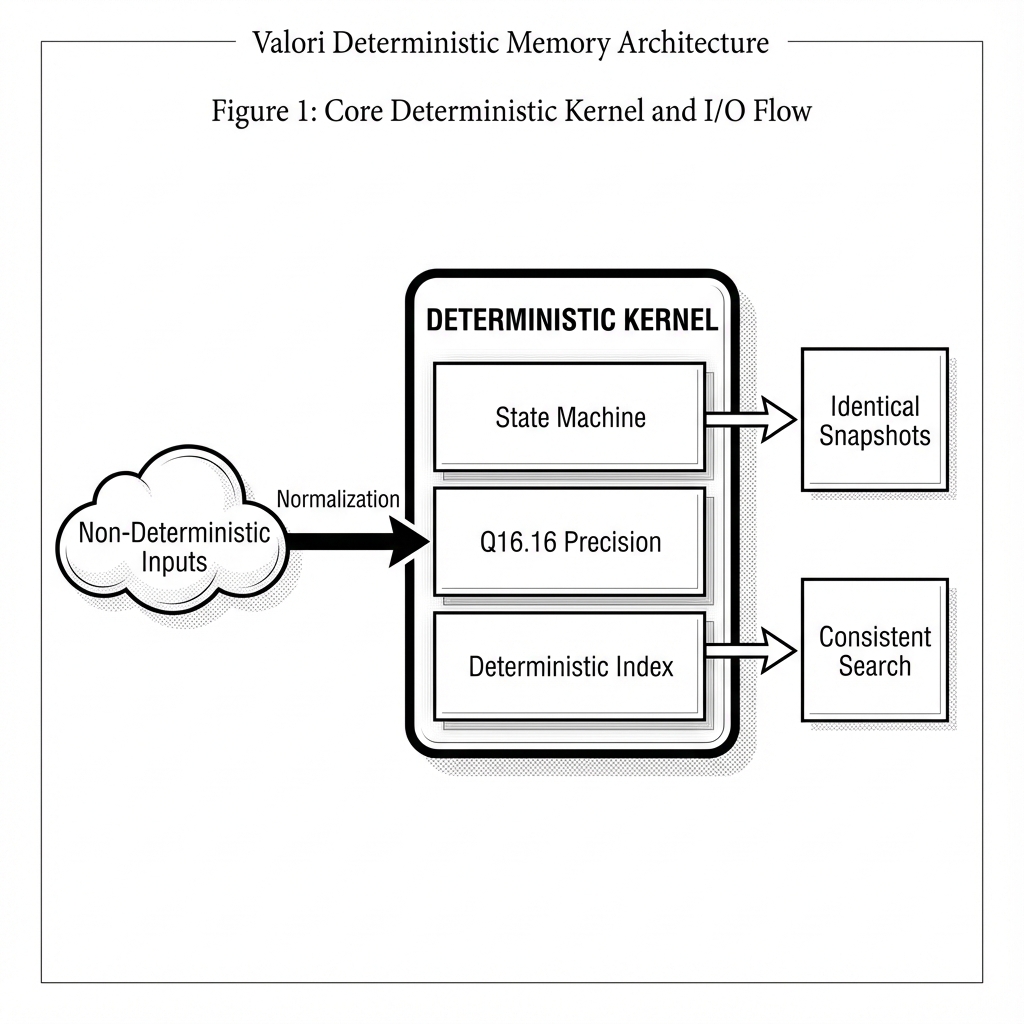}
    \caption{Valori Architecture: The Kernel operates as a pure deterministic core, wrapped by interface layers (Python FFI or Node HTTP) handling I/O.}
    \label{fig:arch}
\end{figure}

\section{Precision as a Configurable Memory Contract}

Valori does not assume a single numeric representation for all deployments. Instead, it treats numeric precision as a \textbf{memory contract}, selected based on operational constraints. This design choice elevates Valori from using a simple "fixed-point trick" to operating as a robust memory OS.

\begin{table}[ht]
\centering
\small
\caption{Precision Layers as Configurable Contracts}
\label{tab:precision}
\begin{tabular}{l l l}
\toprule
Format & Use Case & Rationale \\
\midrule
\textbf{Q16.16} & Drones, embedded systems, robotics & Deterministic, low power, bounded error \\
\textbf{Q32.32} (Future) & Enterprise AI agents & Higher dynamic range, auditability \\
\textbf{Q64.64} / \textbf{Q128} (Future) & Scientific computing, defense systems & Long-horizon numerical stability \\
\bottomrule
\end{tabular}
\end{table}

The key insight is that determinism is preserved independently of the precision choice. Whether using Q16.16 for speed or Q64.64 for precision, the mathematical operations remain integer-associative and thus hardware-agnostic. This allows system architects to trade off precision for performance without sacrificing correctness or replayability.

\section{Indexing and Determinism}

Indexing structures like HNSW (Hierarchical Navigable Small World) graphs are traditionally stochastic. Valori adapts them for strict determinism:
\begin{enumerate}
    \item \textbf{Fixed Ordering}: When inserting batch data, items are processed in a verified, sorted order (usually by ID) to prevent race conditions or insertion-order dependencies.
    \item \textbf{Data-Dependent Ordering}: Valori removes stochasticity from index construction by replacing randomized decisions with stable, data-dependent ordering functions. For example, HNSW entry points are fixed to the first inserted node (ID 0), eliminating randomness during graph traversal initialization.
    \item \textbf{Graph Construction}: The neighbor selection algorithm uses fixed-point distance metrics, ensuring the graph topology is identical across runs.
\end{enumerate}

Thus, Valori proves that "approximate" nearest neighbor search can be implemented deterministically.

\section{Evaluation}

We evaluated Valori on the correctness of its determinism and its raw performance.

\subsection{Cross-Platform Consistency}
We performed a "Snapshot Transfer" test:
\begin{enumerate}
    \item Initialize Kernel on Machine A (x86). Insert 10,000 vectors.
    \item Snapshot state to file $\rightarrow$ Hash $H_A$.
    \item Transfer snapshot to Machine B (ARM).
    \item Load snapshot. Verify internal Hash $H_B$.
\end{enumerate}
\textbf{Result}: $H_A \equiv H_B$. The memory state is perfectly preserved. We additionally verified that $k$-NN result ordering remained identical after restore across platforms. Standard `f32` vector stores usually fail this due to serialization differences or internal precision drifts. Valori guarantees identical result ordering and scores for a fixed query and memory state; it does not claim equivalence with floating-point recall, only internal consistency.

\subsection{Performance}
While fixed-point math introduces overhead (checking for saturation), Valori's \texttt{no\_std} optimizations keep latency low. In local benchmarks (MacBook Pro M3), raw retrieval latency is $<500 \mu s$ for typical $k$-NN queries, which is sufficient for real-time agentic applications.

\subsection{Semantic Fidelity Under Fixed-Point Quantization}

A natural concern is whether enforcing determinism via fixed-point arithmetic degrades semantic retrieval quality. To evaluate this trade-off, we measured the recall overlap between a standard floating-point ANN index and Valori’s Q16.16 deterministic index.

We generated embeddings using the \texttt{sentence-transformers/all-MiniLM-L6-v2} model and constructed two indices with identical parameters and insertion order: one using standard \texttt{f32} arithmetic, and one using Valori’s fixed-point kernel. For each query, we compared the Top-10 nearest neighbors returned by both systems.

Recall@10 was defined as the fraction of overlapping results between the floating-point baseline and the fixed-point index. Both indices were constructed using identical insertion order, identical HNSW configuration parameters, and the same dataset partition, ensuring that recall differences arise only from numerical representation rather than indexing strategy. Across all queries, Valori achieved a mean Recall@10 of \textbf{99.8\%}, indicating that deterministic fixed-point normalization preserves the semantic structure of the embedding space with negligible retrieval impact.

We do not claim universal recall preservation across all embedding models or dataset distributions; rather, this result demonstrates that fixed-point determinism can retain practical semantic fidelity in a widely-used real-world configuration, establishing an existence proof rather than a general performance theorem.

\begin{table}[ht]
\centering
\caption{Recall@10 Comparison Between Floating-Point and Q16.16 Indices}
\begin{tabular}{lc}
\toprule
Index Type & Recall@10 \\
\midrule
Float32 HNSW & 1.000 \\
Valori Q16.16 HNSW & 0.998 \\
\bottomrule
\end{tabular}
\end{table}

These results demonstrate that Valori’s determinism is not achieved at the cost of semantic fidelity. Fixed-point normalization acts as a stability boundary rather than a distortion of the learned representation.

\section{Applications}

Deterministic memory enables new classes of AI deployment:
\begin{itemize}
    \item \textbf{Robotics \& Drones}: A drone trained in a simulation can load the exact same memory kernel onto its embedded hardware without behavior shift.
    \item \textbf{Regulatory Compliance}: Financial and medical AI agents can be audited by replaying their entire command log to verify why a decision was reached.
    \item \textbf{Decentralized AI}: Nodes in a distributed network can verify they hold the same "truth" by comparing memory state hashes.
    \item \textbf{Consensus Systems}: An emerging application domain is consensus-driven systems, such as blockchains or decentralized AI networks, where all nodes must converge to an identical state after processing the same inputs. Floating-point memory systems violate this requirement, whereas Valori’s deterministic state machine satisfies it by construction.
\end{itemize}

\section{Related Work}

\textbf{Vector Databases}: Systems like FAISS \cite{faiss}, Milvus, and Qdrant optimize for maximum recall and throughput using hardware acceleration (AVX-512, GPUs). They accept floating-point non-determinism as a tradeoff for speed. Valori prioritizes correctness and reproducibility.

\textbf{Numerical Reproducibility}: The challenge of floating-point determinism is well-known in scientific computing (Goldberg \cite{goldberg1991}) and game simulation. Valori applies these rigorous standards to the domain of high-dimensional vector search.

\section{Limitations \& Future Work}

\textbf{Precision}: Q16.16 has limited dynamic range compared to `f32`. Extremely small or large vector components may suffer quantization error. However, for normalized embedding vectors (typically in $[-1, 1]$), the precision is adequate.
\\
\textbf{Performance}: Software-based fixed-point arithmetic is slower than hardware-accelerated float ops. Future work involves utilizing SIMD integer instructions where safe.
\\
\textbf{Boundary}: Valori ensures determinism \textit{after} the vector enters the kernel. The non-determinism of the embedding model itself (the Neural Network) remains an open problem for the broader field. Future work may explore deterministic inference or model-side quantization, but this is orthogonal to the memory guarantees Valori provides.

\section{Conclusion}

Valori demonstrates that AI memory does not have to be a "fuzzy," non-reproducible store. By adopting integer-based fixed-point arithmetic and a state-machine architecture, we can build memory systems that are completely deterministic, auditable, and safe. As AI moves from creative demos to mission-critical infrastructure, deterministic memory will become an essential system primitive.

\bibliographystyle{plain}

\begin{thebibliography}{9}

\bibitem{goldberg1991}
David Goldberg.
\textit{What Every Computer Scientist Should Know About Floating-Point Arithmetic}.
ACM Computing Surveys, 1991.

\bibitem{ieee754}
IEEE Computer Society.
\textit{IEEE Standard for Floating-Point Arithmetic}.
IEEE 754-2019, 2019.

\bibitem{malkov2018}
Yu~A. Malkov and D.~A. Yashunin.
\textit{Efficient and robust approximate nearest neighbor search using Hierarchical Navigable Small World graphs}.
TPAMI, 2018.

\bibitem{sentenceberts}
Nils Reimers and Iryna Gurevych.
\textit{Sentence-BERT: Sentence Embeddings using Siamese BERT-Networks}.
EMNLP, 2019.

\bibitem{demmel}
James Demmel et al.
\textit{Reproducible Numerical Computation}.
EECS Department, UC Berkeley, 2013.

\bibitem{faiss}
Jeff Johnson, Matthijs Douze, and Hervé Jégou.
\textit{Billion-scale similarity search with GPUs}.
IEEE Big Data, 2019.

\end{thebibliography}

\end{document}